\pdfoutput=1
\documentclass[acmlarge, screen,nonacm]{acmart}

\usepackage{cleveref}   %
\usepackage{arydshln}
\usepackage{float}
\usepackage{pifont}
\usepackage{wrapfig}
\usepackage{multirow}
\usepackage{tikz}
\usepackage{soul}
\usepackage[breakable, most]{tcolorbox}
\usepackage{geometry}
\usepackage{svg}
\usepackage{amsmath}
\usepackage{soul}
\usepackage{makecell}
\usepackage{multirow}
\usepackage{xcolor}
\usepackage{caption}
\usepackage{subcaption}
\usepackage{enumitem}

\usepackage{layouts}
\usepackage{card}

\usepackage{colorprofiles}
\usepackage[a-1b,mathxmp]{pdfx}[2018/12/22]

\usetikzlibrary{arrows.meta, positioning, fit, shapes.geometric}

\makeatletter
\renewcommand\p@subfigure{\thefigure\,}

\DeclareCaptionLabelFormat{mysublabelfmt}{\alph{sub\@captype}}
\makeatother

\setcopyright{none}
\renewcommand\footnotetextcopyrightpermission[1]{}
\AtBeginDocument{%
  \providecommand\BibTeX{{%
    \normalfont B\kern-0.5em{\scshape i\kern-0.25em b}\kern-0.8em\TeX}}}

\makeatletter
\let\@authorsaddresses\@empty
\makeatother

\title{Enhancing Paraphrase Type Generation: The Impact of DPO and RLHF Evaluated with Human-Ranked Data}

\author{Christopher Lee Luebbers}
\hypersetup{
  pdfkeywords={paraphrasing, DPO, RLHF, natural language processing, human evaluation},
  pdfsubject={Master's Thesis}
}

\affiliation{%
  \institution{University of Göttingen}
  \city{Göttingen}
  \country{Germany}
}

\begin{document}

\renewcommand{\shortauthors}{Christopher L. Lübbers}

\begin{abstract}

\section*{Abstract}

Paraphrasing re-expresses meaning to improve text simplification, machine translation, and question-answering. 
Specific paraphrase types enable accurate semantic analysis and robust language models. 
However, current paraphrase-type generation methods do not fully align with human preferences because they rely on automated metrics and limited human-annotated training data. 
These shortcomings obscure crucial aspects of semantic fidelity and linguistic transformations defined by paraphrase types.

We address this gap by leveraging a human-ranked paraphrase-type dataset and integrating Direct Preference Optimization (DPO) to align model outputs directly with human judgments. 
DPO-based training increases paraphrase-type generation accuracy by 3 percentage points over a supervised baseline and raises human preference ratings by 7 percentage points. 
Our newly created human-annotated dataset supports more rigorous future evaluations. 
We also provide a paraphrase-type detection model that achieves F1 scores of 0.91 for addition/deletion, 0.78 for same polarity substitution and 0.70 for punctuation changes.  

These findings demonstrate that preference data and DPO training produce more reliable, semantically accurate paraphrases, enabling downstream applications such as improved summarization and more robust question-answering. 
The PTD model surpasses automated metrics and provides a more reliable framework for evaluating paraphrase quality. 
This approach advances paraphrase-type research toward richer, user-aligned language generation and establishes a stronger foundation for future evaluations grounded in human-centric criteria.
\end{abstract}

\maketitle

\section{Introduction}

Paraphrasing transforms a sentence's form while preserving its core meaning \citep{paraphrase}, enabling more effective machine translation, summarization, and question-answering. 
For example, 'The scientist explained the results clearly' can be paraphrased as 'The scientist clarified the results.'
By generating faithful yet varied reformulations, models become more robust, better able to adapt to shifting domains, and more attuned to user expectations in interactive systems \citep{li_paraphrase_2018, iyyer2018adversarialexamplegenerationsyntactically}. 

Within paraphrasing research, atomic paraphrase types (APT) define fine-grained linguistic transformations \citep{vila_is_2014}. 
Addition ('The scientist \textit{thoroughly} explained the results clearly'), or same polarity substitution, where a synonym preserves meaning ('The scientist \textit{described} the results clearly') exemplify transformations that specify precise linguistic modifications rather than broad similarity. 
Building on APT, Paraphrase Type Generation (PTG), and Paraphrase Type Detection (PTD) \citep{wahle_paraphrase_2023} drive the field forward by focusing model outputs and analyses on specific transformations. 
Despite the importance of PTG and PTD, models rarely produce or detect paraphrases that consistently convey intended meanings while following subtle linguistic shifts. 

This deficiency arises from a fundamental gap: a lack of high-quality, human-ranked data that specifies which paraphrases users find preferable and why. 
Widely used metrics like BLEU and ROUGE \citep{papineni-etal-2002-bleu, lin-2004-rouge} focus on lexical overlap rather than semantic fidelity and cannot detect if paraphrasing was applied correctly \citep{shen2022evaluation}. 
Creating human-ranked paraphrase datasets requires extensive effort \citep{chen-dolan-2011-collecting}, and the scarcity of such data restricts progress. 
Without sufficient human annotation that ranks paraphrases on subtle qualities, models struggle to internalize these preferences, especially for complex transformations \citep{meier2024humanunderstandingparaphrasetypes, wang2022deep}.

\begin{figure*}[htbp]
    \centering
    \includegraphics{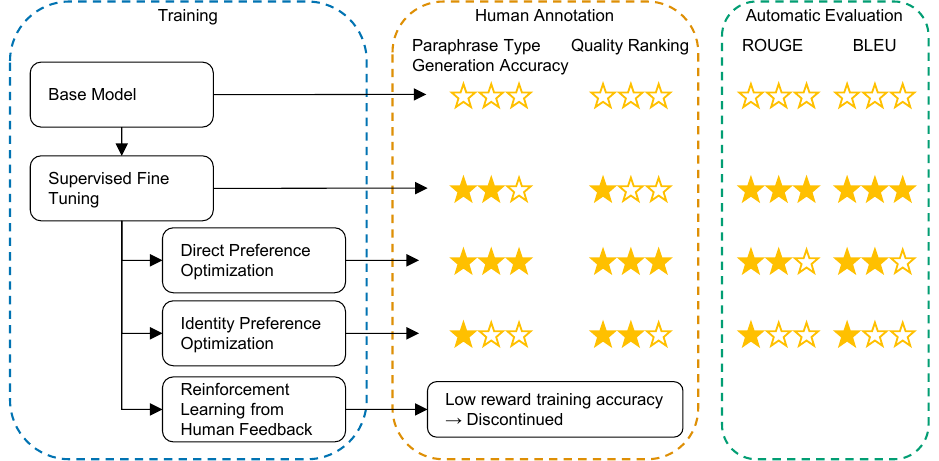}
    \caption{Comparison of direct preference optimization (DPO), identity preference optimization (IPO), and reinforcement learning from human feedback (RLHF) against supervised fine-tuning and a baseline. 
    We used Annotated Paraphrase TYpes (APTY) and Extended Typology Paraphrase Corpus (ETPC) datasets.  
    The diagram outlines paraphrase-type generation steps, evaluation metrics, and the incorporation of human feedback. 
    Stars indicate increasing alignment quality, and the discontinued pathway marks where RLHF performed suboptimally}
    \label{fig:teaser}
\end{figure*}

This research addresses that gap by leveraging human-ranked data to guide paraphrase generation and detection. 
We\footnote{'We' is used to align with academic conventions, reflecting collective research efforts, though this thesis is the work of a sole author.} employ Reinforcement Learning from Human Feedback (RLHF), Direct Preference Optimization (DPO), and Identity Preference Optimization (IPO) \citep{azar2023generaltheoreticalparadigmunderstand} to generate paraphrases that incorporate target APT-defined transformations more faithfully. 
Working with the Annotated Paraphrase TYpes (APTY) dataset \citep{meier2024humanunderstandingparaphrasetypes}, we guide models to produce paraphrases that exhibit the desired APT-defined transformations and reflect authentic human preferences. 
Figure~\ref{fig:teaser} illustrates the approach, comparing optimization techniques, showing data flow, and incorporating human feedback. 
Our evaluations confirm that DPO-based models yield more accurate and user-preferred paraphrases, reveal the shortcomings of traditional evaluation metrics, and highlight the value of human-centric training for improved language understanding.

Key contributions include:
\begin{enumerate}    
    \item Enhanced paraphrase-type generation accuracy(\cref{sec:accuracy}): 
    DPO training on APTY increases human-annotated accuracy by 3~\% over a supervised baseline, aligning outputs with nuanced linguistic transformations.
    \item Improved user-aligned quality (\cref{sec:human_preferences}): 
    Human evaluators favor these improved outputs 7~\% more than baseline paraphrases, underscoring enhanced semantic fidelity and stylistic appropriateness.
    \item A new human-ranked dataset: 
    The dataset we produce enables a more rigorous, fine-grained evaluation of paraphrase quality and paves the way for future research. 
    \item Exposing metric limitations (\cref{sec:evaluation_comparison}): 
    Weak correlations (Spearman's $r<0.3$) between automated metrics and human rankings motivate the development of richer evaluation frameworks.
    \item Improved paraphrase-type detection (\cref{sec:ptd}): 
    Our PTD model achieves F1 scores of 0.91 on addition/deletion, 0.78 on same polarity substitution, and 0.70 for punctuation changes, enabling more granular assessments.
    \item Improved reasoning (\cref{sec:generalizability}): 
    PTG boosts multistep soft reasoning (MuSR) task performance by 38~\%, demonstrating broader benefits for language generation and reasoning tasks.
\end{enumerate}
These advances provide a framework that better aligns PTG and PTD with human needs, benefiting translation, summarization, and conversational AI applications.

\section{Related Work}

Paraphrase generation aims to produce meaning-preserving reformulations of text, yet state-of-the-art methods struggle to reflect human-defined preferences for nuanced linguistic transformations. 
Multiple studies introduced typologies and evaluation methods, yet most approaches still fail to incorporate human preferences at a granular level.

\citet{vila_is_2014} established the APT taxonomy, identifying transformations like addition, deletion, and substitution. 
The APT typology provides foundational insight into linguistic variation in paraphrases, surpassing generic similarity-based approaches.
Building on APT, \citet{wahle_paraphrase_2023} defined PTG and PTD tasks.  
They demonstrated that models could generate and identify these APT-defined transformations but found that outputs did not align with human judgments. 

This mismatch underscored the need to incorporate user-centric criteria into model training. 
To address this shortfall, \citet{meier2024humanunderstandingparaphrasetypes} introduced the APTY-ranked dataset, which included human rankings of paraphrases. 
They discovered that large models like ChatGPT performed well on simpler paraphrase types but struggled with complex transformations. 
Their findings revealed the critical role of direct human feedback in guiding models toward more nuanced, preference-aligned outputs. 
This reveals a critical limitation: models generate typologically diverse paraphrases that fail to align with human preferences. 
These alignment gaps motivate our research, aiming to incorporate human rankings into model training effectively.

Another major challenge involves evaluating paraphrase quality. 
Traditional measures such as BLEU \citep{papineni-etal-2002-bleu} and ROUGE \citep{lin-2004-rouge} rely on lexical overlap, failing to capture semantic fidelity or detect actual paraphrasing. 
\citet{shen2022evaluation} found that reference-free metrics surpass reference-based ones.
They propose ParaScore, a metric that explicitly models lexical divergence but weakly correlates with human judgments.
Similarly, \citet{zhou_paraphrase_2022} show that under-representing paraphrase types and using single references reduce evaluation reliability. 
\citet{oh-etal-2023-evaluation} showed that multiple pseudo-references improved approximations of human preferences but still lacked an entirely human-centered framework. 

Pre-trained models such as BERT \citep{devlin2019bertpretrainingdeepbidirectional} and GPT \citep{Radford2018ImprovingLU} improved paraphrase fluency, as \citet{wang2022deep} found, yet they optimized surface-level similarities rather than deep semantic fidelity. 
\citet{huang-etal-2023-paraamr} introduced ParaAMR to enhance paraphrase syntactic diversity but did not incorporate human preference signals. 
Although reinforcement learning emerged to encode human evaluations, its application to fine-grained paraphrase types remains limited. 
Prior efforts focused on fluency and syntax \citep{li_paraphrase_2018}, not human-judged transformations. 
\citet{rafailov_direct_2023} proposed DPO to align model outputs with human rankings without complex reward models. 
\citep{azar2023generaltheoreticalparadigmunderstand} modified the loss function of DPO to counter overfitting. 
DPO and IPO allow optimizing for user preferences at finer levels, addressing PTG gaps.

These gaps, including typology without human alignment, unreliable metrics, and the absence of fine-grained optimization, underscore the need for a new framework. 
Unlike previous works, our approach combines DPO with APTs to generate paraphrases aligned with nuanced human preferences. 
This framework aligns with human preferences and supplies robust evaluation and fine-grained optimization techniques for practical applications.

\section{Methodology}

This methodology addresses the stated research questions by proposing a formal problem setting, detailing datasets, presenting step-by-step procedures, and describing evaluation protocols. 
We integrate human-ranked data, DPO, and a PTD model to enhance the generation and evaluation of paraphrase types.

\subsection{Research Questions and Problem Definition}

We consider three key research questions:
(RQ1) Can integrating human-ranked data with DPO improve PTG accuracy and user preference compared to baseline and reward-based approaches?
(RQ2) Can a PTD model verify the presence and correctness of fine-grained transformations, offering a more nuanced evaluation than traditional metrics?
(RQ3) Do improvements in PTG and PTD, driven by human-centric optimization, enhance performance on broader NLP tasks that require semantic fidelity?

We use PTG and PTD definitions from \citet{wahle_paraphrase_2023}. 
We define PTG as follows: Given an original sentence $x$ and a target transformation $l_i \in L$, where $L$ is a set of APT, produce a paraphrase $\Tilde{x}$ that preserves the meaning of $x$ while applying transformation $l_i$.

For PTD, given a pair $(x,\Tilde{x})$, the task is to identify which APT categories $l_i$ were applied to $x$. PTD enables reference-free evaluation of transformations and provides deeper insight into paraphrase quality.

\subsection{Paraphrase Type Generation}

\begin{figure*}[htbp]
    \centering
    \includegraphics{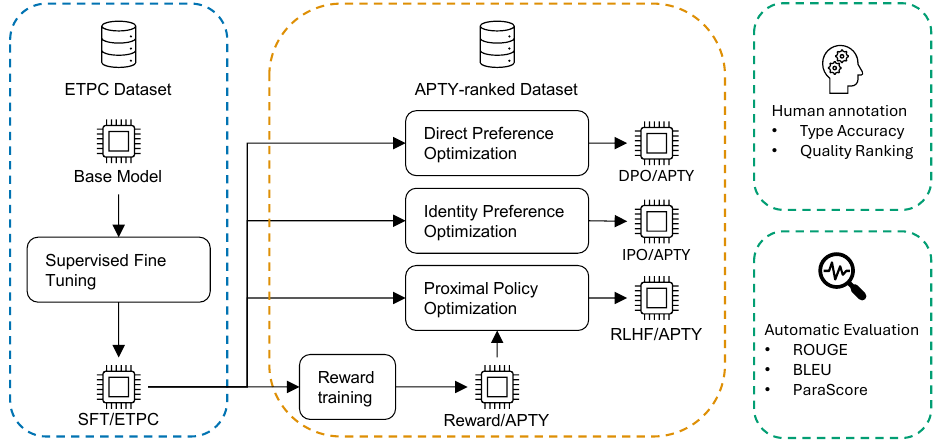}
    \caption{Paraphrase Type Generation (PTG) workflow. 
    The model is first fine-tuned on the Extended Typology Paraphrase Corpus (ETPC) to learn atomic paraphrase transformations, then further refined with the Annotated Paraphrase Type (APTY) dataset using reward training, direct preference optimization (DPO), and identity preference optimization (IPO). 
    Final evaluation employs ROUGE, BLEU, and human annotations to confirm accurate, preference-aligned paraphrases.}
    \label{fig:ptg_pipeline}
\end{figure*}

The PTG pipeline involved three steps, as illustrated in \cref{fig:ptg_pipeline}. 
First, we fine-tuned the base model with supervised fine-tuning (SFT) on the Extended Typology Paraphrase Corpus (ETPC) \citep{kovatchev-etal-2018-etpc}.  
Second, we further trained the fine-tuned model on APTY-ranked using reward training, DPO, and IPO. 
Finally, we evaluated the resulting models with ROUGE, BLEU, and human annotations on ETPC examples.

\subsubsection{Datasets} 

We use datasets in the English language. 
ETPC contains 5,800 paraphrase examples categorized by APT. 
With extensive coverage and prior validation \citep{wahle_paraphrase_2023}, ETPC ensures a solid typological foundation during SFT and prepares the model for integrating human preferences. 
We employed a 70-30 training-test split and used the same prompt as \citet{wahle_paraphrase_2023}:
$$
\label{eq:prompt}
\begin{aligned}
    &\text{Given the following sentence, generate a paraphrase with the following type.} \\
    &\text{Sentence: ['original']} \\
    &\text{Paraphrase Types: ['APT'].} \\
    &\text{Answer: }
\end{aligned}
$$

The preprocessed APTY-ranked dataset, consisting of 333 human-ranked paraphrase examples, provides fine-grained human rankings, addressing ETPC's limitations. 
APTY's ranked preferences introduce a human-centric dimension, representing a key innovation over prior work that relied on static reference-based comparisons. 
As the only dataset with human-ranked paraphrase types, APTY-ranked offers a novel perspective on the quality and diversity of paraphrase generation. 
We preprocessed APTY-ranked by fixing encoding issues, stripping whitespace, normalizing punctuation, and applying an 80-20 split stratified by paraphrase types. 
We used the same prompt as in ETPC (\cref{eq:prompt}). 
APTY-ranked complements ETPC by incorporating human judgments into training.

\subsubsection{Models and Training}

We chose LLaMA-3.1-8B \citep{dubey2024llama} and LLaMA-2-7B \citep{touvron2023llama2openfoundation} models for open-source availability. 
These models scale for large-scale NLP tasks, offer parameter efficiency, and enable resource-efficient training for diverse NLP applications. 
We employed Low-Rank Adaptation (LoRA) \citep{hu2021loralowrankadaptationlarge} layers on Llama models to optimize resource usage. 
This approach reduces computation and allows focused adaptation, supporting reproducibility and broader adoption. 
However, LoRA fine-tuning may limit generalization for intricate reasoning \citep{shuttleworth2024loravsfinetuningillusion}.
We also included BART \citep{lewis2019bartdenoisingsequencetosequencepretraining} for ParaScore comparisons since ParaScore supports limited models.

The training involved multiple stages. 

\begin{figure}[htbp]
    \centering
    \includegraphics[width=\linewidth]{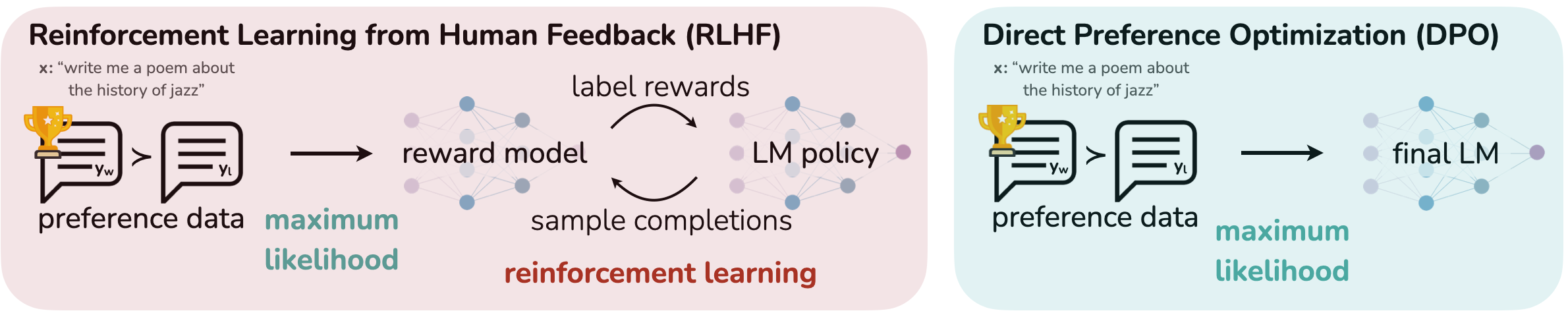}
    \caption{Direct Preference Optimization (DPO) framework, adapted from \citet{rafailov_direct_2023}. 
    DPO aligns model outputs with human rankings by directly optimizing for preferred responses, eliminating the need for a separate reward model or reinforcement learning. 
    This streamlined approach results in paraphrases more closely matching user-defined quality criteria.}
    \label{fig:dpo}
\end{figure}

\begin{enumerate}
    \item SFT on ETPC: We trained BART on ETPC to establish a baseline understanding of paraphrase types. 
    \citet{wahle_paraphrase_2023} provided a Llama-3.1-8B model fine-tuned on ETPC.
    \item After SFT, we refined the models using APTY-ranked. 
    We explored three optimization techniques:
    \begin{itemize}
        \item Reward Training: Uses a reward model to guide PTG toward higher-quality outputs, depending on a predefined reward structure. 
        This model would serve in Proximal Policy Optimization \citep{schulman2017proximalpolicyoptimizationalgorithms}, a standard approach for transformer-based RLHF. 
        \item DPO aligns outputs with human rankings, bypassing a separate reward model. 
        \cref{fig:dpo} shows an overview of the method. 
        We manually tuned hyperparameters to maximize reward margins, representing the mean difference between chosen and rejected rewards. 
        Besides maximizing reward margins, we balanced accuracy and stable loss stability. 
        A trial table is available in the repository \citep{repo}.
        \item IPO counters overfitting by adjusting the loss function of DPO. 
        For the IPO model, we also tuned hyperparameters to maximize reward margins and maintain accuracy and stable loss \citep{repo}. 
    \end{itemize}
\end{enumerate}

\subsubsection{Evaluation}

We conducted a thorough evaluation combining automatic metrics and human judgments. 
This dual evaluation bridges gaps between automatic metrics and human preferences. 
We computed ROUGE-1, ROUGE-2, ROUGE-L, and BLEU from ETPC references. 
Due to tool constraints, we applied ParaScore only to BART outputs. 
While these metrics provide baselines, their known misalignment with human judgment demands more human-centric approaches.

Two human reviewers evaluated outputs for semantic fidelity and type adherence, providing insights into user alignment and complementing metrics. 
The reviewers were one master's student in Applied Data Science and one doctoral student. 
Multiple annotators mitigated subjectivity and bias. 
This step validates whether metric improvements correspond to human preferences. 
The reviewers received a sentence with an APT and model-generated paraphrases. 
Annotators decided if the specified APT was correctly and exclusively applied. 
They ranked valid paraphrases on a scale of 1 (best) to 4 (worst). 
If the generated sentence was not paraphrased or the specified type was absent, annotators assigned ranked 4. 
All valid paraphrases were ranked from 1 downward.  
We used Pearson correlation \citep{pearson} to evaluate the correlation between rank annotations and metrics. 
Our annotations schema assigns '4' to all incorrect paraphrases, creating a rigid valid-invalid boundary. 
This discontinuity mismatches the discrete human annotation scale and the continuous ROUGE/BLEU scores, reducing correlation. 
We applied a logistic transformation to smooth ranks and align them with continuous metrics.  
We used $$Transformed Score = \frac{1}{1+\exp(Original Score - 2.5)}$$ with 2.5 as the anotation midpoint. 
This transformation introduces score continuity.  

We compared broader performance on Open LLM Leaderboard v2 \citep{open-llm-leaderboard-v2} tasks: Instruction-Following Evaluation (IFEval) \citep{zhou2023instructionfollowingevaluationlargelanguage}, Big Bench Hard (BBH) \citep{suzgun2022challengingbigbenchtaskschainofthought}, MATH Lvl 5 \citep{hendrycks2021measuringmathematicalproblemsolving}, Graduate-Level Google-Proof Q\&A Benchmark (GPQA) \citep{rein2023gpqagraduatelevelgoogleproofqa}, Multistep Soft Reasoning (MuSR) \citep{sprague2024musrtestinglimitschainofthought}, Massive Multitask Language Understanding - Professional (MMLU-Pro) \citep{wang2024mmluprorobustchallengingmultitask}.

\subsection{Paraphrase Type Detection}

\begin{figure*}[htbp]
    \centering
    \includegraphics{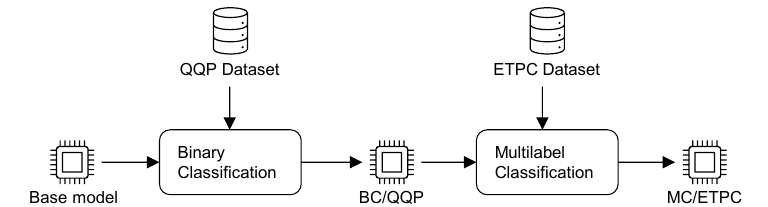}
    \caption{Paraphrase Type Detection pipeline. 
    Initial training on the Quora Question Pairs (QQP) dataset teaches paraphrase recognition, followed by fine-tuning on a filtered ETPC subset to identify specific atomic transformations. 
    Stratified sampling and weighted losses address class imbalances, ensuring robust detection of diverse paraphrase types.}
    \label{fig:ptd_pipeline}
\end{figure*}

Evaluations often rely on references, which may be limited or unavailable. 
PTD verifies which transformations a given paraphrase applies, enabling reference-free evaluation and complementing PTG.
If a model excels at generating certain paraphrase types, can a detection model confirm these transformations? 
Figure \ref{fig:ptd_pipeline} shows the PTD pipeline.  

\subsubsection{Datasets}

For PTD, we first trained a model on Quora Question Pairs (QQP) \citep{qqp} to learn paraphrase vs. non-paraphrase distinctions. 
QQP provides a widely used benchmark, offering a solid foundation before fine-tuning on more nuanced data. 
We used the train split, which contains 364k rows, for training, and the validation split, which contains 40.4k rows, for testing. 

Next, we fine-tuned on a filtered ETPC set, focusing on the top 10 paraphrase types: addition/deletion, change of order, derivational changes, inflectional changes, punctuation changes, same polarity substitution (contextual), semantic-based, spelling changes, subordination and nesting changes, synthetic/analytic substitution (\cref{tab:dataset_etpc}). 
This prevents performance loss for rare types.  
It aligns with APTY's focus on the top 10 types. 
We tracked each example's paraphrase types, enabling type-based organization and count maintenance. 
A balanced split prioritized rare paraphrase types, ensuring representation in training and testing. 
We used an 80-20 training-testing split, preserving the distribution of paraphrase types. 
Handling duplicates proved critical, as examples could belong to multiple paraphrase types. 
We used sets to avoid repeated multiple subset assignments.  
A weighted BCEWithLogitsLoss addresses imbalances, giving rare types receive proper attention.

\subsubsection{Model and Training}

We selected DeBERTa \citep{he2021debertadecodingenhancedbertdisentangled} for its advanced architecture and strong classification performance. 
Our two-stage fine-tuning, binary classification on QQP, and multilabel classification on ETPC are innovative approaches to bridging basic paraphrase recognition with fine-grained type detection. 
We optimized class weights, learning rate, weight decay, and batch size with Optuna to guarantee robust results. 
Meticulous tuning ensures accurate type detection, supporting targeted model evaluations.

\subsubsection{Evaluation}

We used Macro-F1 scores on ETPC to measure balanced performance across all paraphrase types. 
Macro-F1 prevents the dominance of common types. 
Weighted loss and type-stratified sampling address type imbalances, providing equitable detection performance. 
A validation set and systematic hyperparameter tuning ensured reliable improvements. 

\section{Results}

We tested the hypothesis that incorporating human-ranked preferences via DPO improves PTG accuracy for specific transformations, increases human satisfaction with the outputs, reveals limitations of current automated metrics, and enhances performance in complex reasoning tasks. 
We conducted a series of experiments to validate each component of this hypothesis, using multiple models, human annotations, and both established and novel evaluation approaches. 

\subsection{Effect of DPO on Paraphrase Type Generation Accuracy}\label{sec:accuracy}

We analyzed the accuracy of trained models in generating specific APT. 
Preliminary Llama-2-7B experiments showed a significant accuracy increase. 
While Llama-2-7B and SFT models achieved 12~\% accuracy, DPO reached 35~\% (\cref{fig:llama2_apt_accuracy}). 
We continued to evaluate Llama-3.1-8B (baseline), SFT/ETPC, DPO/APTY, and IPO/APTY. 
Because the reward model accuracy was only 49~\%, we discontinued RLHF, highlighting its limitations. 
Each model generated single-type paraphrases on 260 base sentences. 
This produced 1040 human-annotated paraphrases covering the top 10 APTs. 
The base sentences are available on Github\footnote{\url{https://github.com/worta/generate_apt_paraphrases}}. 
We defined accuracy as the proportion of generated paraphrases correctly exhibiting the target transformation without semantic loss. 

\begin{figure*}[htbp]
    \centering
    \includegraphics{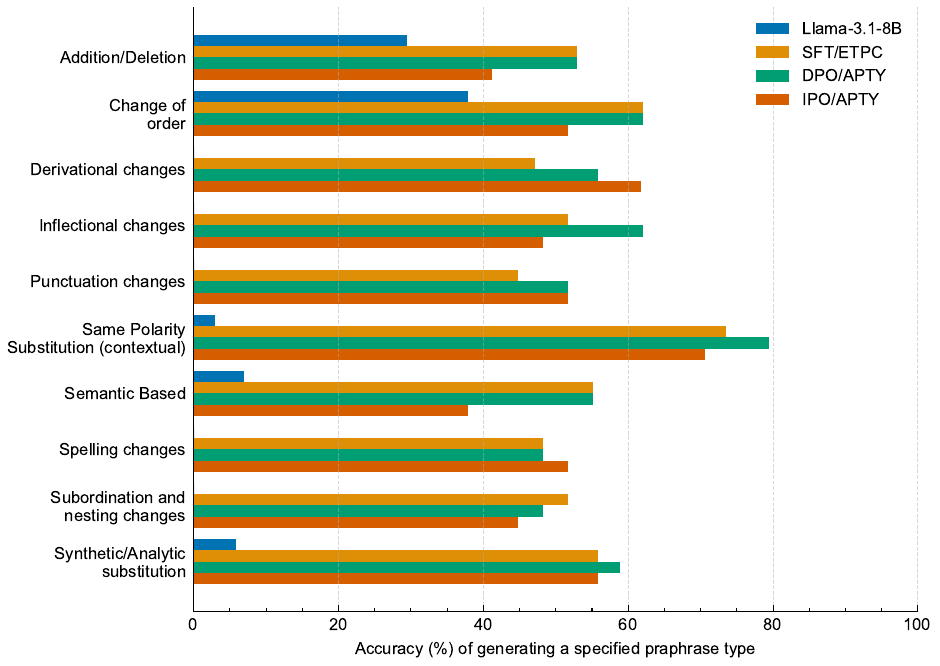}
    \caption{Accuracy of paraphrase-type generation across methods. 
    The DPO/APTY model achieves 57~\% accuracy, surpassing both the baseline Llama-3.1-8B (8~\%) and a supervised fine-tuning (SFT/ETPC) approach. 
    Accuracy reflects the proportion of paraphrases that apply the targeted transformation without semantic loss, validated by human annotations. 
    Certain transformations, such as punctuation and word-order changes, remain challenging for all models.
    The base model Llama-3.1-8B could not generate a correct paraphrase for some paraphrase types, resulting in missing bars.}
    \label{fig:llama3_apt_accuracy}
\end{figure*}

DPO/APTY model achieved 57~\% mean accuracy (SD=9), outperforming Llama-3.1-8B (8~\%, SD=14), SFT/ETPC (54~\%, SD=8) and IPO/APTY (52~\%, SD=10). 
A one-way ANOVA confirmed significant differences ($F(3,1036)=49.4, p<10^{-12}$). 
Inter-annotator agreement was substantial (Cohen's $\kappa=0.69$ \citep{Landis1977TheMO}), ensuring reliable human assessments. 

DPO's human-aligned approach improves handling complex types (\cref{fig:llama3_apt_accuracy}). 
Despite many examples, challenging types (semantic transformations, inflectional changes, word-order alterations, punctuation changes) remained difficult. 
DPO/APTY's gains show that human-ranked data yields better paraphrase type alignment than baselines and alternatives. 
While accuracy shows DPO's technical skill, human preference analysis reveals practical effectiveness. 

\subsection{Human Preferences for DPO-Generated Paraphrases}\label{sec:human_preferences}

We next asked whether these accuracy gains translate into outputs that humans favor. 
Human annotators ranked 260 paraphrase sets, each containing 1 paraphrase by each model, from best (1) to worst (4), assigning the worst rating to invalid or incorrectly transformed paraphrases. 

\begin{figure}[htbp]
    \centering
    \includegraphics{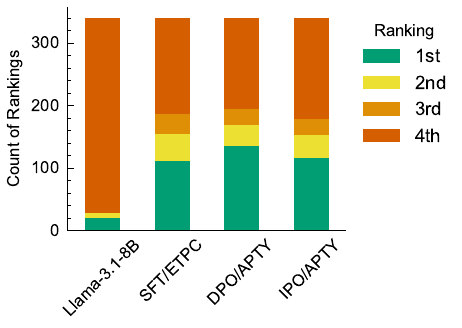}
    \caption{Human preference rankings of model-generated paraphrases. 
    The DPO/APTY model produces 49~\% top-ranked paraphrases (rank 1) compared to 33~\% for the baseline SFT/ETPC. 
    Rankings (1–4) consider both correctness and adherence to the specified paraphrase type. 
    The DPO/APTY model’s consistently superior rankings highlight its closer alignment with human judgments.}
    \label{fig:llama3_ranking}
\end{figure}

DPO/APTY achieved 40~\% top-rank vs. baseline's 6~\% and SFT/ETPC's 33~\% (\cref{fig:llama3_ranking}), $\chi^{2}(9)=231.9, p<10^{-44}$. 
\cref{tab:percentage_rankings} details the rankings. 
Krippendorff's \citep{krippendorffContentAnalysisIntroduction2024} $\alpha=0.63$ indicated moderate inter-annotator agreement. 

DPO's advantage persisted across categories, including complex types (e.g., punctuation, word-order changes). 
DPO/APTY aligns outputs with human preferences, improving quality across types. 
These findings validate our approach and support critiques of traditional metrics \citep{shen2022evaluation}. 
By prioritizing semantic accuracy and nuanced transformations, DPO aligns with \citet{vila_is_2014} and \citet{wahle_paraphrase_2023}, advancing human-aligned paraphrase generation.  
Although humans favor DPO, we must assess automated metrics for alignment.  

\subsection{Automated Metrics vs. Human Preferences}\label{sec:evaluation_comparison}

\begin{figure*}[htbp]
    \centering
    \includegraphics{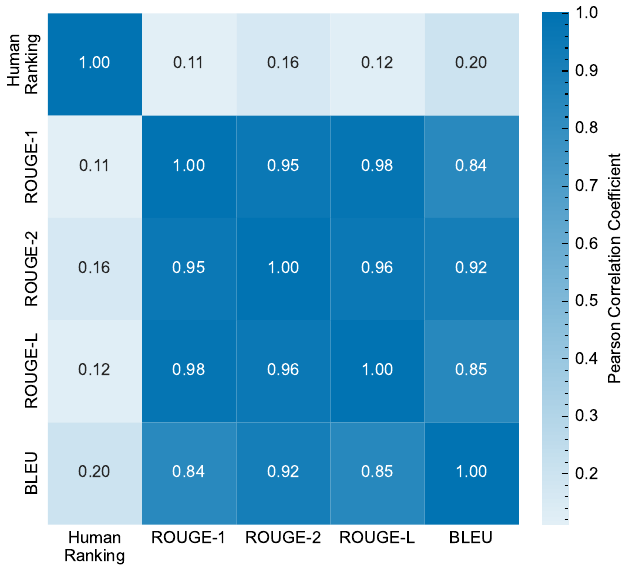}
    \caption{Correlation between automated metrics and human quality judgments. 
    For 120 paraphrases from four models, Pearson correlations of ROUGE and BLEU with human rankings remain below 0.3, indicating weak alignment. 
    Human rankings were transformed to a logistic scale for compatibility. 
    These results underscore the need for more human-centered evaluation methods.}
    \label{fig:llama3_correlation_matrix}
\end{figure*}

We correlated human rankings with ROUGE and BLEU scores to assess whether standard metrics recognize these improvements.
Annotators ranked 30 random paraphrase sets with base sentences and references taken from ETPC. 
As shown in Figure~\ref{fig:llama3_correlation_matrix}, Pearson correlations fell below 0.3 ($p<10^{-10}$).

Despite human preference for DPO/APTY, even advanced metrics like ParaScore showed weak alignment (\cref{tab:scores}). 

\begin{table*}[htbp]
  \centering
  \caption{Automated metric scores (ROUGE, BLEU, ParaScore) for an illustrative paraphrase from SFT/ETPC and DPO/APTY fine-tuned BART models. 
  Despite the SFT/ETPC output not constituting a valid paraphrase, automated metrics remain high due to lexical overlap with the reference. 
  This example demonstrates the limitations of standard metrics in capturing genuine paraphrase quality.}
  \begin{tabular}{lll}
    \toprule
    Metric / Base Sentence & SFT/ETPC  & DPO/APTY\\
    \midrule
    Those reports were denied & Those reports were denied & Those reports, however,\\
    by the interior minister, & by Prince Nayef, the & were denied by the minister, \\
    Prince Nayef. & interim minister of education. & Prince Nayef \\
    ROUGE-1 & 0.47 & 0.42  \\
    ROUGE-2 & 0.09 & 0.11  \\
    ROUGE-L & 0.28 & 0.31  \\
    BLEU    & 0.04 & 0.05  \\
    ParaScore Base  & 0.87 & 0.84  \\
    ParaScore Free  & 0.86 & 0.83  \\
    \bottomrule
  \end{tabular}  
  \label{tab:scores}
\end{table*}

These results show standard metrics fail as proxies for human perception \citep{shen2022evaluation, oh-etal-2023-evaluation}. 
Our findings confirm that DPO/APTY models produce human-preferred paraphrases but not necessarily higher lexical scores. 
Automatic metrics fail to detect whether the generated sentence qualifies as a paraphrase. 
Improved metrics aligned with human judgment are essential. 
We developed a PTD model for a more nuanced evaluation. 

\subsection{Paraphrase Type Detection with a Novel Model}\label{sec:ptd}

To move beyond traditional metrics, we developed a PTD model that classifies paraphrase transformations without relying on reference sentences. 

\begin{figure*}[htbp]
    \centering
    \includegraphics{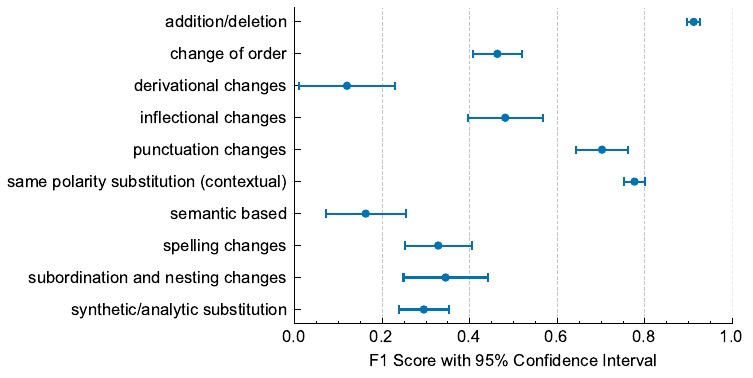}
    \caption{F1 scores for paraphrase type detection (PTD) across ten atomic transformations. 
    The PTD model excels at simpler types (Addition/Deletion: F1=0.91; Same Polarity Substitution: F1=0.78) but struggles with complex semantic shifts. 
    Scores are derived from 1,040 human-annotated examples, highlighting variability in detection difficulty.}
    \label{fig:deberta_ptd_f1}
\end{figure*}

Trained on the top 10 APTs, this model achieved a weighted F1 of 0.71, performing best on simpler transformations (e.g., addition/deletion: F1=0.91, 95\% CI [0.90, 0.93])) and less well on complex semantic shifts (\cref{fig:deberta_ptd_f1}). 
Details are noted in \cref{tab:f1_scores}. 
We evalauted the model on ETPC, where each example can have multiple APT. 
We also evaluated how this PTD model agrees with human annotation on the 260 single APT transformation sets we used for accuracy and ranking (\cref{fig:deberta_ptd_human}). 

Although not a complete solution, the PTD model provides more granular insights, enabling fine-grained evaluation that traditional metrics fail to offer. 
This supports the central claim by demonstrating the value of tools aligned with the conceptual framework of APTs for evaluating paraphrase quality.

\subsection{Impact of DPO on Broader NLP Tasks}\label{sec:generalizability}

\begin{figure*}[htbp]
    \centering
    \includegraphics{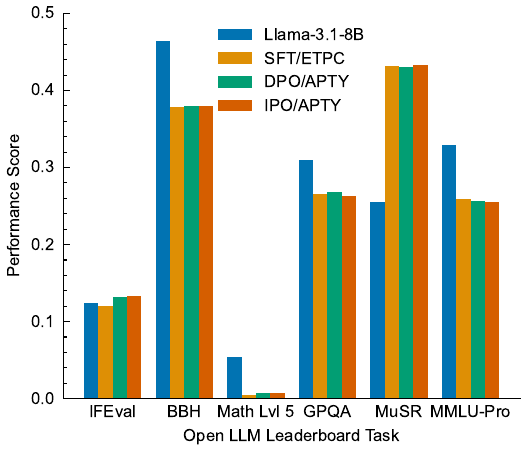}
    \caption{Model performance on various NLP benchmarks, including Multistep Soft Reasoning (MuSR) and Big Bench Hard (BBH). 
    The DPO/APTY model notably improves MuSR performance, which requires complex reasoning, though it shows minor declines on other tasks. 
    The bars represent model scores, illustrating how human preference alignment influences performance across multiple domains.}
    \label{fig:llama3_llmboard}
\end{figure*}

Finally, we evaluated whether improvements driven by human-ranked paraphrase-type data extend to broader tasks. 
We tested all models on Open LLM Leaderboard v2 benchmarks. 
While DPO/APTY showed marginal declines on some tasks (\cref{fig:llama3_llmboard}), it improved MuSR team allocation performance by up to 38~\% (\cref{tab:musr_scores}. 
The PTG models ranked 7th among all 85 Llama-3.1-8B models on the Open LLM Leaderboard MuSR ranking (2024-12-20). 

\begin{table*}[htbp]
\caption{Detailed Multistep Soft Reasoning (MuSR) task results. 
Scores indicate the model’s ability to handle multi-step reasoning processes and integrate contextual information, reflecting performance on complex inference tasks.
Optimized models excel on team allocation tasks.}
\label{tab:musr_scores}
\begin{tabular}{llll}
\toprule
 Model & Murder mysteries & Object placements & Team allocation \\
\midrule
Llama-3.1-8B & 0.54 & 0.35 & 0.25 \\
SFT/ETPC & 0.51 & 0.35 & \textbf{0.43} \\
DPO/APTY & 0.52 & 0.34 & \textbf{0.42} \\
IPO/APTY & 0.51 & 0.35 & \textbf{0.43} \\
\bottomrule
\end{tabular}
\end{table*}

This result indicates that human-centered PTG training benefits paraphrasing and certain aspects of reasoning, bolstering the overarching hypothesis that human preference alignment enhances language models across multiple dimensions.

\section{Conclusion}

This study addressed a critical gap in PTG by applying DPO and using human-ranked data to produce paraphrases that align with nuanced human preferences. 
While earlier work often relied on automated metrics and limited human feedback, we now see that integrating direct human judgments improves transformation-specific accuracy by 3~\% and increases user preference by 16~\%. 
These results answer the central research question: DPO-trained models yield more reliable, context-sensitive paraphrases than those relying solely on baseline methods.

Our PTD model provides a rigorous evaluation approach. 
This model verified the presence of specific transformations and surpassed conventional metrics such as ROUGE and BLEU. 
Although the detector excelled at identifying more straightforward changes, it struggled with subtler linguistic shifts, reflecting the complexity of capturing nuanced semantics.
This outcome suggests that future methods should refine detection approaches and incorporate richer semantic cues.

However, certain limitations remain. 
We discontinued RLHF due to low reward model accuracy, which highlights the limitations of this application. 
A reward model is trained on a classification task. 
The differences in human preferences in APTY might be too subtle for the model. 
The APTY dataset’s limited size and subjective annotations reduced generalizability, and subtle semantic transformations and complex word-order alterations remained difficult. 
Future work should expand training corpora, incorporate multiple languages and modalities, and develop evaluation metrics that more effectively capture semantic, contextual, and stylistic fidelity.
While LoRA improved computational efficiency, it may have reduced complex paraphrasing performance.  
Addressing these challenges requires systematically evaluating DPO with different architectures or fine-tuning strategies. 
Standardized frameworks blending multiple metrics weighted by human correlation could better reflect linguistic aspects. 

This study provides a foundation for producing semantically rich, human-preferred paraphrases.  
Integrating human judgments in training and evaluation brings NLP closer to user preferences.  
These advances offer a blueprint for bridging the gap between technical optimization and real-world language demands. 

The code is available on GitHub \citep{repo} and the trained models are on Huggingface \citep{huggingface}. 
\begin{acks}
Dominik Meier and Dr. Terry Lima Ruas from the Chair for Scientific Information Analytics (Prof. Dr. Bela Gipp) at the University of Göttingen provided the idea for this project. 
Jan Philip Wahle implemented supervised fine-tuning of Llama models on the ETPC dataset as part of the EMNLP'23 paper "Paraphrase Types for Generation and Detection"\citep{wahle_paraphrase_2023}. 
The doctoral student Dominik Meier supervised the project at the University of Göttingen. 

I am deeply grateful to my daughters and my wife for their endless encouragement, love, and patience. Their tireless support and belief in me carried me through late nights and tough moments, making this thesis possible.
\end{acks}

\bibliographystyle{ACM-Reference-Format}
\bibliography{references}

\appendix
\section{Technical Information}

A requirements file is available on Github \citep{repo}. 
The table (\cref{tab:library_versions}) and requirements file ensure reproducibility and transparency for exact replication.

\begin{table}[htbp]
    \centering
    \caption{Python libraries and corresponding versions used in the experimental environment. 
    Specifying exact versions ensures reproducibility and facilitates independent validation of the reported results.}
    \begin{tabular}{ll}
        \toprule
        Library & \textbf{Version} \\
        \midrule
        python & 3.11.10 \\ 
        NumPy & 2.1.3 \\ 
        pandas & 2.2.3 \\
        PyTorch & 2.5.1 \\ 
        torchvision & 0.19.1 \\
        torchaudio & 2.4.1 \\ 
        cuda & 12.1 \\
        scikit-learn & 1.5.2 \\
        spacy & 3.7.5 \\ 
        scipy & 1.14.1 \\ 
        seaborn & 0.13.2 \\ 
        matplotlib & 3.9.2 \\
        transformers & 4.46.2 \\ 
        datasets & 3.1.0 \\ 
        trl & 0.12.0 \\
        accelerate & 1.1.1 \\ 
        bitsandbytes & 0.44.1 \\
        peft & 0.13.2 \\
        rouge & 1.0.1 \\     
        nltk & 3.9.1 \\        
        \bottomrule
    \end{tabular}
    \label{tab:library_versions}
\end{table}

\section{Datasets}\label{sec:dataset}

ETPC refines Microsoft Research Paraphrase Corpus \citep{dolan-brockett-2005-automatically} by annotating it with the Extended Paraphrase Typology (EPT). 
This annotation introduces granular distinctions between paraphrase and non-paraphrase, including contextual, habitual, and negation relations. 
ETPC enhances MRPC utility by enabling detailed evaluation and error analysis and supporting tasks like semantic similarity and entailment. 
We used the Oct 2, 2023 version.
\cref{tab:dataset_etpc} shows APT frequencies. 

The APTY dataset \citep{meier2024humanunderstandingparaphrasetypes} extends ETPC with human-ranked preferences for APT-based paraphrases. 
APTY includes APTY-base (correctness and errors) and APTY-ranked (human preferences). 
The dataset columns include 'original', 'chosen', and 'rejected'.  
We used the July 8, 2024 version. 
\cref{tab:dataset_etpc} shows APT frequencies after preprocessing. 

\begin{table}[htbp]
\caption{Frequency counts of paraphrase types in the Extended Typology Paraphrase Corpus (ETPC) and the Annotated Paraphrase Type (APTY) dataset. 
In ETPC, multiple instances of a paraphrase type can occur within a single sentence, explaining total vs. unique counts. 
In APTY, multiple chosen and rejected examples may relate to the same original sentence.}
\label{tab:dataset_etpc}
\begin{tabular}{lrrrr}
\toprule
 & ETPC total & ETPC unique & APTY total & APTY unique\\
\midrule
Addition/Deletion & 5722 & 2988 & 40 & 9 \\
Change of format & 240 & 207 &&\\
Change of order & 860 & 766 & 60 & 10\\
Converse substitution & 43 & 42 &&\\
Coordination changes & 48 & 47 &&\\
Derivational Changes & 187 & 181 & 22 & 6\\
Diathesis alternation & 162 & 161 &&\\
Direct/indirect style alternations & 66 & 66 \\
Ellipsis & 66 & 64 &&\\
Entailment & 81 & 81 &&\\
Identity & 3878 & 3870 &&\\
Inflectional Changes & 613 & 544 & 12 & 7\\
Modal Verb Changes & 184 & 180 &&\\
Negation switching & 20 & 20 &&\\
Non-paraphrase & 842 & 605 &&\\
Opposite polarity substitution (contextual) & 15 & 12 \\
Opposite polarity substitution (habitual) & 4 & 4 \\
Punctuation changes & 833 & 748 & 28 & 9\\
Same Polarity Substitution (contextual) & 4173 & 2511 & 56 & 10 \\
Same Polarity Substitution (habitual) & 840 & 681 &&\\
Same Polarity Substitution (named ent.) & 536 & 448 &&\\
Semantic-based & 337 & 328 & 59 & 10\\
Spelling changes & 636 & 534 & 31 & 9\\
Subordination and nesting changes & 473 & 448 & 18 & 5\\
Syntax/discourse structure changes & 308 & 305 &&\\
Synthetic/Analytic Substitution & 897 & 806 & 7 & 5\\
\bottomrule
\end{tabular}
\end{table}

\section{Models}

We trained models with parameter configurations enhancing paraphrase generation. 
Key parameters include:

\begin{itemize}
    \item LoRA target modules for Llama-3-8B: up\_proj, down\_proj, k\_proj, o\_proj, v\_proj, gate\_proj, and q\_proj
    \item DPO parameters: learning rate = 1e-6, weight decay = 4e-1, $\beta = 0.2$, max\_grad\_norm = 200, lr\_scheduler = cosine
    \item IPO parameters; warmup ratio = 0.2, weight decay = 0.02, learning rate = 5e-6, $\beta = 0.2$, lr\_scheduler = 'reduce learning rate on plateau'
\end{itemize}

\section{Evaluation}

We annotated generated paraphrases with Label Studio 1.10.0 available at \url{https://labelstud.io/}. 
Our labeling instructions: 
$$
\begin{aligned}
    &\text{Only evaluate the first paraphrase.} \\
    &\text{All incorrect paraphrases are ranked "wrong".} \\
    &\text{All correct paraphrases are ranked from "best" downwards.}
\end{aligned}
$$

\begin{figure*}[htbp]
    \centering
    \includegraphics[width=\linewidth]{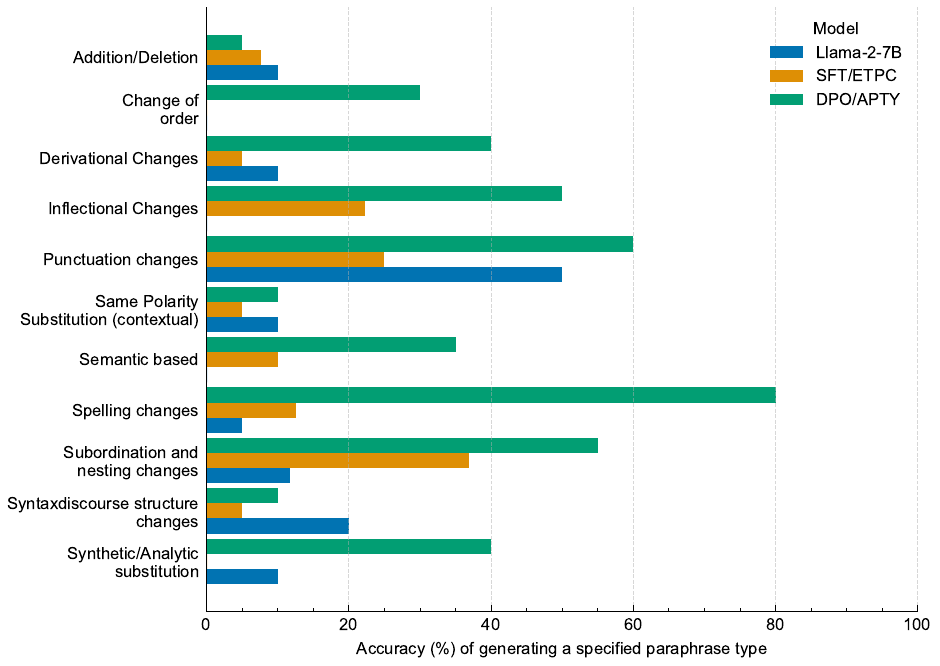}
    \caption{Accuracy of paraphrase-type generation by Llama-2-7B and a DPO/APTY model, based on human annotations. 
    While DPO/APTY outperforms the baseline overall, challenging transformations (e.g., 'change of order', 'derivational changes') remain difficult for both models.}
    \label{fig:llama2_apt_accuracy}
\end{figure*}

We conducted preliminary DPO optimization with Llama-2-7B.  
Mean accuracy rose from 12~\% (base, SFT/ETPC) to 35~\% (DPO/APTY). 
Accuracy by APT is shown in \cref{fig:llama2_apt_accuracy}.

\begin{table}[htbp]
\caption{Distribution of human rankings for paraphrase quality: The table shows each model's human rankings (1 = best, 4 = worst). 
DPO/APTY achieves the highest top-rank proportion; Llama-3.1-8B mostly ranks 4. 
A Chi-Square test (92.34, p=5.5e-16) confirms DPO/APTY's superiority.}
\label{tab:percentage_rankings}
\begin{tabular}{lrrrr}
\toprule
 & 1 & 2 & 3 & 4 \\
\midrule
Llama-3.1-8B & 6 & 2 & 0 & 91 \\
SFT/ETPC & 33 & 13 & 9 & 45 \\
DPO/APTY & 40 & 10 & 8 & 43 \\
IPO/APTY & 34 & 11 & 8 & 47 \\
\bottomrule
\end{tabular}
\end{table}

\begin{table*}
\caption{F1 scores for the paraphrase type detection model across multiple transformation categories. 
The model excels at Addition/Deletion (F1=0.91), Same Polarity Substitution (0.77), and Punctuation (0.70), but achieves lower accuracy for more complex transformations, indicating varied detection difficulty.}
\label{tab:f1_scores}
\begin{tabular}{lrrrr}
\toprule
Class & F1 Score & CI Lower & CI Upper & Support \\
\midrule
Addition/Deletion & \textbf{0.91} & 0.90 & 0.93 & 1327 \\
Change of order & 0.46 & 0.41 & 0.52 & 303 \\
Derivational Changes & 0.12 & 0.01 & 0.23 & 33 \\
Inflectional Changes & 0.48 & 0.39 & 0.57 & 130 \\
Punctuation Changes & \textbf{0.70} & 0.64 & 0.76 & 223 \\
Same Polarity Substitution (contextual) & \textbf{0.78} & 0.75 & 0.80 & 1113 \\
Semantic-Based & 0.16 & 0.07 & 0.25 & 63 \\
Spelling Changes & 0.33 & 0.25 & 0.40 & 144 \\
Subordination and Nesting Changes & 0.34 & 0.25 & 0.44 & 94 \\
Synthetic/Analytic substitution & 0.29 & 0.24 & 0.35 & 248 \\
\bottomrule
\end{tabular}
\end{table*}

\begin{figure*}[htbp]
    \centering
    \includegraphics[width=\linewidth]{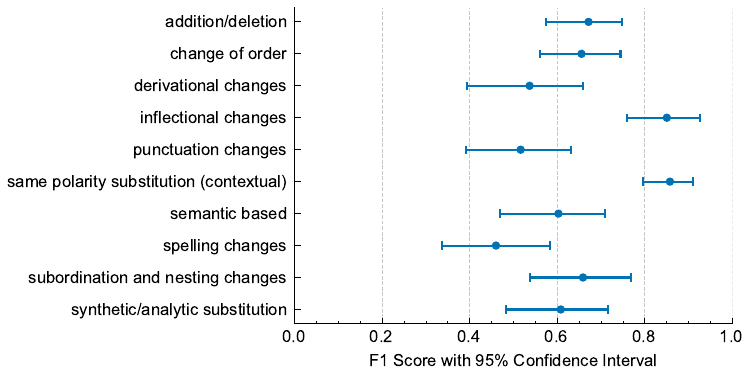}
    \caption{F1 scores for agreement of paraphrase type detection (PTD) with human annotators.  
    The PTD model scores better, because PTG was performed on single transformation.}
    \label{fig:deberta_ptd_human}
\end{figure*}

\clearpage
{\sffamily
    \centering
    \tcbset{colback=white!10!white}
    \begin{tcolorbox}[
        title={\large \textbf{AI Usage Card for \textit{Enhancing Paraphrase Type Generation: The Impact of DPO and RLHF Evaluated with Human-Ranked Data}}\hfill \makebox{\qrcode[height=1cm]{https://ai-cards.org/}}},
        breakable,
        boxrule=0.7pt,
        width=.8\paperwidth,
        center,
        skin=bicolor,
        segmentation empty,
        before lower={\footnotesize{AI Usage Card v1.0 \hfill \url{https://ai-cards.org} \hfill \hfill \href{https://ai-cards.org/whitepaper.pdf}{PDF} | \href{https://ai-cards.org/whitepaper.bib}{BibTeX} | \href{https://ai-cards.org/whitepaper.xml}{XML} | \href{https://ai-cards.org/whitepaper.json}{JSON} | \href{https://ai-cards.org/whitepaper.csv}{CSV}}},
        halign lower=center,
        collower=white,
        colbacklower=tcbcolframe]
        
    \footnotesize{
        \begin{longtable}{p{.15\paperwidth} p{.275\paperwidth} p{.275\paperwidth}}
             {\color{LightBlue} \MakeUppercase{Correspondence(s)}} \newline  Christopher L. Lübbers
             & {\color{LightBlue} \MakeUppercase{Contact(s)}} \newline c.luebbers@stud.uni-goettingen.de 
             & {\color{LightBlue} \MakeUppercase{Affiliation(s)}} \newline University of Göttingen, Chair for Scientific Information Analytics \\\\
             & {\color{LightBlue} \MakeUppercase{Project Name}} \newline Enhancing Paraphrase Type Generation: The Impact of DPO and RLHF Evaluated with Human-Ranked Data & {\color{LightBlue} \MakeUppercase{Key Application(s)}} \newline Natural Language Processing, Paraphrase Generation \\\\
             {\color{LightBlue} \MakeUppercase{Model(s)}} \newline ChatGPT \newline ChatGPT \newline ChatGPT \newline Grammarly
             & {\color{LightBlue} \MakeUppercase{Date(s) Used}} \newline 2024/06/24 - 2024/12/20 \newline  2024/09/12 - 2024/12/04 \newline 2024/12/05 - 2024/12/20 \newline 2024/06/24 - 2024/12/20 
             & {\color{LightBlue} \MakeUppercase{Version(s)}} \newline 4o \newline o1-preview \newline o1 \newline \\\\
             \cmidrule{2-3}\\
             
             {\color{LightBlue} \MakeUppercase{Ideation}} \newline ChatGPT o1-preview, o1 
             & {\color{LightBlue} \MakeUppercase{Improving existing ideas}} \newline Reviewed and refined the research proposal, identifying areas for improvement.\\\\
             
             {\color{LightBlue} \MakeUppercase{Literature Review}} \newline ChatGPT 4o
             & {\color{LightBlue} \MakeUppercase{Finding literature}} \newline Expanded the list of known literature by identifying additional relevant sources.\\\\
               
             {\color{LightBlue} \MakeUppercase{Methodology}} \newline ChatGPT o1-preview, o1 
             & {\color{LightBlue} \MakeUppercase{Finding iterative optimizations}} \newline Enhanced the proposed methodology by addressing potential gaps and increasing rigor. \\\\
                          
             {\color{LightBlue} \MakeUppercase{Writing}} \newline ChatGPT 4o, Grammarly
             & {\color{LightBlue} \MakeUppercase{Generating new text based on instructions}} \newline Transformed bullet points into complete, coherent sentences using ChatGPT.
             & {\color{LightBlue} \MakeUppercase{Assisting in improving own content}} \newline Grammarly refined written content for correctness, clarity, and engagement. \\\\
             & {\color{LightBlue} \MakeUppercase{Paraphrasing related work}} \newline  Summarized related works partially with ChatGPT's assistance. \\\\
             
             {\color{LightBlue} \MakeUppercase{Presentation}} \newline ChatGPT 4o
             & {\color{LightBlue} \MakeUppercase{Generating new artifacts}} \newline Created initial versions of figures and tables, which were later manually improved. \\\\
                         
             {\color{LightBlue} \MakeUppercase{Coding}} \newline ChatGPT 4o
             & {\color{LightBlue} \MakeUppercase{Generating new code based on descriptions or existing code}} \newline Collaboratively developed code in a pair programming setup.
             & {\color{LightBlue} \MakeUppercase{Refactoring and optimizing existing code}} \newline Debugged and optimized code for functionality and efficiency.\\\\
             \cmidrule{2-3}\\
             
             {\color{LightBlue} \MakeUppercase{Ethics}}
             & {\color{LightBlue} \MakeUppercase{What are the implications of using AI for this project?}} \newline AI accelerated analysis and improved clarity for writing.
             & {\color{LightBlue} \MakeUppercase{What steps are we taking to mitigate errors of AI for this project?}} \newline Rigorous manual review, cross-validation, and expert verification ensure accuracy and reliability. \\\\
             & {\color{LightBlue} \MakeUppercase{What steps are we taking to minimize the chance of harm or inappropriate use of AI for this project?}} \newline Sentences for paraphrase generation were carefully chosen manually and annotated by experienced researchers.
             & {\color{LightBlue} \MakeUppercase{The corresponding authors verify and agree with the modifications or generations of their used AI-generated content}} \newline All authors reviewed and approved all AI-generated or AI-modified content.  \\
        \end{longtable}
     }
    \tcblower
    \end{tcolorbox}
}

\end{document}